\documentclass[pmlr]{jmlr}
\usepackage[utf8]{inputenc}

\usepackage{xcolor}
\usepackage{fnpct}
\usepackage{graphicx}
\usepackage{booktabs}
\usepackage{multirow}
\usepackage{amsmath, amsfonts, bm}

\usepackage{mathtools}

\jmlrvolume{179}
\jmlryear{2022}
\jmlrworkshop{Conformal and Probabilistic Prediction and Applications}
\jmlrproceedings{PMLR}{Proceedings of Machine Learning Research}

\title[Calibration of NLU Models with Venn--ABERS Predictors]{Calibration of Natural Language Understanding Models with Venn--ABERS Predictors}

\author{
    \Name{Patrizio Giovannotti} \Email{patrizio.giovannotti.2019@live.rhul.ac.uk}\\
    \addr{Royal Holloway, University of London, Egham, Surrey, UK}
}

\editor{Ulf Johansson, Henrik Boström, Khuong An Nguyen, Zhiyuan Luo and Lars Carlsson}

\begin{document}

\maketitle

\begin{abstract}
Transformers, currently the state-of-the-art in natural language understanding (NLU) tasks, are prone to generate uncalibrated predictions or extreme probabilities, making the process of taking different decisions based on their output relatively difficult. In this paper we propose to build several inductive Venn--ABERS predictors (IVAP), which are guaranteed to be well calibrated under minimal assumptions, based on a selection of pre-trained transformers. We test their performance over a set of diverse NLU tasks and show that they are capable of producing well-calibrated probabilistic predictions that are uniformly spread over the [0,1] interval -- all while retaining the original model's predictive accuracy.
\end{abstract}


\section{Introduction}
Natural language understanding (NLU) systems are rapidly becoming an essential part of many commercial products. A core element in the architecture of any conversational agent, NLU tasks are also behind machine translation, question answering and text classification tasks like sentiment analysis, hate speech detection and fake news detection. Such a shift from the academic realm to a more product-oriented environment was possible because of the recent advancements in natural language processing: massive new datasets, better computing resources, the introduction of pre-training and the transformer resulted in a dramatic performance jump across many tasks.

However, once embedded in a product, any mistake made by the NLU system can result in sub-optimal service, misleading results or even undermine the whole usability of the product (case in point, a faulty speech-to-text engine). In order to mitigate the effects of wrong predictions, NLU systems need the ability to reliably assess their own uncertainty. Mitigation strategies include asking the user for feedback, or even refusing to produce an output, should the model judge its prediction to be too uncertain.

As an example, let us consider a traditional spam detector. Depending on its uncertainty over an incoming message $m$, a spam detector may act in several ways: if the calculated probability of $m$ being spam is $\mathbb{P}(Y=\text{spam}\mid m)=p$, then
\begin{itemize}
    \item If $p>k_1$, where $k_1$ is a suitable threshold, say 0.95, send the message to the recycle bin or spam folder
    \item If $k_2<p<k_1$, where $k_2<k_1$ is a lower threshold, ask the user to double check if the message is actually spam
    \item If $p<k_2$ do not take any action.
\end{itemize}
In order to produce such a behaviour, a model must not only be accurate in detecting spam: it needs to be able to estimate realistic probabilities tailored to each different message, that is, it needs to be \textit{well-calibrated} and \textit{sharp}.
A well-calibrated model is able to output probabilities that match the observed frequencies of the predicted labels. For example, out of all predictions with an estimated probability of 0.85, exactly 85\% of them must be correct predictions.

However, good calibration alone may not be enough: for a test set where spam and not spam are equally frequent, a classifier assigning probability $p=0.50$ to each prediction would indeed be well calibrated, although hardly useful. A sharper model could divide the examples in more than one ``category'' and assign probabilities accordingly. In the example above, at least 3 categories would be desirable.

In this work we propose to use inductive Venn--ABERS predictors (IVAPs) to build well-calibrated, sharp NLU models. IVAPs have the property of being perfectly calibrated under minimal assumptions. We apply IVAPs to several types of transformer models \citep{vaswani2017attention}, a class of pre-trained neural architectures that are currently the state-of-the-art in NLU. Transformers, however, are not guaranteed to be well calibrated and have a strong tendency to output ``extreme'' probabilities (close to 0 or 1) -- hence unable to distinguish any example that lies in between (see \figureref{fig:roberta-qqp}). We show that transformer-based IVAPs are well calibrated and tend to produce a uniform distribution of probability scores: they are sharp in that sense. We test the performance of IVAPs against a range of different NLU tasks which, given the nature of IVAPs, we restrict to the \textit{binary} case.

The code to reproduce our results is available on GitHub.\footnote{\url{https://github.com/patpizio/vennabers-for-nlu}} This includes a link to an interactive Colab notebook. 

\begin{figure}%[htbp]
\floatconts
    {fig:roberta-qqp}% label
    {\caption{Reliability bubble chart for a RoBERTa model trained on the QQP dataset. The vast majority of predictions are concentrated in only two output probabilities. The few exceptions are grossly uncalibrated.}}% caption command
    {\includegraphics[width=0.6\textwidth]{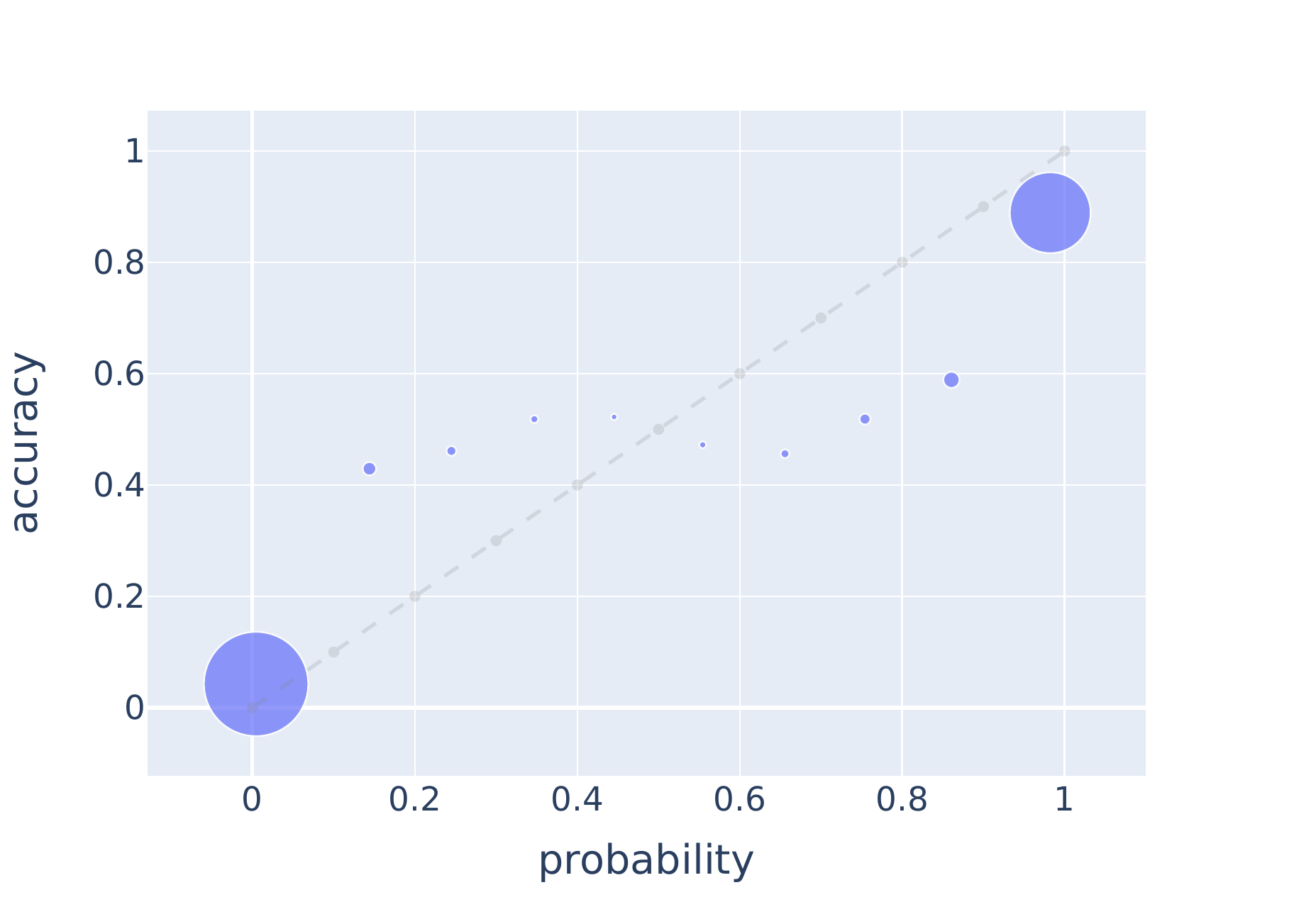}}
\end{figure}

\section{Venn--ABERS predictors}
Venn--ABERS predictors \citep{Vovk2014VennAbersP} are a special case of Venn predictors \citep{vovk2005algorithmic}, a class of probabilistic predictors guaranteed to be valid under the sole assumption of the training examples being exchangeable. Like all Venn predictors, they need two adjustments to hold their validity guarantee: i) they output multiple probability distributions over the labels -- one for each possible label -- and ii) their validity property is restricted to perfect calibration. This is because it can be proven that it is impossible to build a valid probabilistic predictor, in the general sense \citep{gammerman1998}.

Formally, calibration could be defined as follows: let the random variable $Y\in\{0,1\}$ model the label predicted by a binary classifier. Let $P\in[0,1]$ be the confidence associated to the same prediction. $P$ is perfectly calibrated if
\[
\mathbb{E}(Y|P)=P
\]
almost surely. 

Venn--ABERS predictors are binary predictors and output a pair of probabilities $(p_0, p_1)$ for each test example $(x,y)$. The former is the probability of $y=1$ should the true label be 0, while the latter is the probability of $y=1$ should the true label be 1: only one of the two is the valid prediction, but we don't know which one (as we don't know $y$). Because we always have $p_0<p_1$, the pair $(p_0,p_1)$ can be interpreted as the lower and upper probabilities, respectively, of a certain prediction.  Depending on the test example, $p_0$ and $p_1$ may be more or less different in magnitude, although they are usually close to each other. A large gap between $p_0$ and $p_1$ signifies low confidence in the probability estimation -- something traditional probabilistic predictors are not able to provide. For practical reasons however, it is often useful to have one probability estimate per test example. A reasonable way to combine the two numbers, as explained in \citet{Vovk2014VennAbersP}, is to calculate the probability which minimizes the regret for the log loss function:
$$
p=\frac{p_1}{1-p_0+p_1}\:.
$$

In this work we will be using the \textit{inductive} variant of VAPs (IVAP), which was proposed as a computationally lighter version of VAPs in \citet{vovk2015large}. This is our only choice as the traditional VAP needs to be retrained for each test example, something absolutely infeasible given the average training time of a transformer model.

IVAPs can be created as follows. Suppose we have a binary classification problem and a \textit{scoring algorithm}, i.e. any ML algorithm that can issue any confidence score for each prediction -- in our case, a transformer model. The general procedure to fit an IVAP is the following:
\begin{enumerate}
    \item Divide the training set made of examples $(x, y)$  in a \textit{proper training set} of size $l$ and a \textit{calibration set} of size $k$
    \item Train the transformer on the proper training set
    \item Obtain the scores $z_1,\dots,z_k$ for the objects $x_1,\dots,x_k$ in the calibration set
    \item For a test example $x$, calculate its score $z$. Fit one \textit{isotonic regression} on the set $(z_1,y_1),\dots,(z_k,y_k),(z,0)$, then another one on the set $(z_1,y_1),\dots,(z_k,y_k),(z,1)$ so to obtain two functions $f_0$ and $f_1$.
    \item IVAP outputs the multiprobability $(p_0,p_1)=(f_0(z),f_1(z))$
\end{enumerate}
Isotonic regression is a nonparametric form of regression that fits a step-wise, non-decreasing function to a set of examples (see \citealp{zadrozny2002transforming}).
IVAPs still require for the isotonic regression to be re-calculated for each test example, for each label. Fortunately, \citet{vovk2015large} designed an optimised version that requires a single pre-calculation step ($\mathcal{O}(l\log l)$), then performs an efficient $\mathcal{O}(\log l)$ evaluation step for every test example. We use a Python implementation released by Paolo Toccaceli.\footnote{\url{https://github.com/ptocca/VennABERS}}

\section{Related work}
Given the recent developments in the state-of-the-art, the analysis of calibration in NLU tasks \citep{nguyen-oconnor-2015-posterior} is gradually turning into the analysis of calibration of transformer models. While \citet{pmlr-v70-guo17a} warned about the tendency of ``deep'' models to produce miscalibrated predictions, \citet{desai-durrett-2020-calibration} and \citet{minderer2021revisiting} showed that recent transformer architectures in particular could be well-calibrated out of the box. However, other studies like the one of \citet{10.1162/tacl_a_00407} reported rather poor calibration scores for transformer models on generative question answering datasets. In general, all transformer models seem to benefit from additional calibration steps, and there is no substantial research about their sharpness.

Several recent research contributions focused on building valid predictors to estimate uncertainty in NLU tasks. Mostly based on traditional conformal prediction, models have been built for text classification \citep{pmlr-v105-paisios19a}, sentiment analysis \citep{pmlr-v128-maltoudoglou20a}, paraphrase detection \citep{pmlr-v152-giovannotti21a} and part-of-speech tagging and text infilling \citep{dey2021conformal}. Conformal prediction was also applied to relation extraction \citep{pmlr-v139-fisch21a} and fact verification \citep{fisch2021efficient}.

Venn--ABERS predictors have been successfully applied in different fields, such as drug discovery \citep{buendia2019accurate}, compound activity prediction \citep{toccaceli2016excape}, adversarial manipulation detection \citep{peck2020detecting} and log anomaly detection \citep{pan2020anomaly}. To the best of our knowledge, we are the first to apply Venn–ABERS prediction to modelling uncertainty in NLU.

% \begin{itemize}
%     \item Calibrating transformers
%     \item Other calibration techniques
%     \begin{itemize}
%         \item Isotonic regression?
%         \item Temperature scaling
%         \item Platt scaling?
%         \item Histogram binning?
%         \item Label smoothing?
%     \end{itemize}
% \end{itemize}

\section{Experiments}
In this section we provide the details of our experimental process: datasets used, transformer models considered and performance metrics adopted. Because many of these datasets are used in ongoing competitions, their test set labels may be hidden. For these datasets we select the labelled examples and shuffle them into two new training / development sets (see the details in Appendix \ref{sec:apx_training}).

\subsection{Datasets} \label{datasets}
We tried to include binary datasets that could test our models across different NLU abilities or tasks.
\paragraph{Quora Question Pairs (QQP)} is a large dataset for \textit{paraphrase detection}: the task of determining if two sentences are semantically equivalent. Each example is a pair of questions taken from those asked on the Quora website. QQP is currently realeased as part of a Kaggle competition.\footnote{\url{https://www.kaggle.com/c/quora-question-pairs}}
\paragraph{Stanford Sentiment Treebank (SST)} is a sentiment scoring dataset: each data item is a film review extract labelled with a real number between 0 and 1 that indicates its level of positive sentiment. In our work we transform it into a binary dataset by rounding each label to the nearest integer. SST was introduced by \citet{socher-etal-2013-recursive}.
\paragraph{Corpus of Linguistic Acceptability (CoLA)} consists of English acceptability judgements drawn from books and journal articles on linguistic theory. Each example is a sequence of words annotated with whether it is a grammatical English sentence. CoLA was introduced in \citet{warstadt2018neural} and is currently used in the GLUE public benchmark.
\paragraph{Boolean Questions (BoolQ)} is a question answering dataset for yes/no questions which are naturally occurring -- they are generated in unprompted and unconstrained settings. Each example is a triplet of (question, passage, answer). BoolQ was introduced in \citet{clark-etal-2019-boolq} and is currently used in the SuperGLUE public benchmark.

\subsection{Pre-trained models}
Transformer models are rarely trained from scratch, as they are designed to take advantage of large amounts of data and computational resources. Instead, a common practice in ML research is to re-use such pre-trained models by training them again on smaller datasets, which may even model an NLU task different from the one seen at pre-training step. This process, known as fine-tuning, has proven to be beneficial across many benchmarks and allows for the use of powerful models without excessive demands in terms of needed resources.

Following \citet{bowman2021combating}'s encouragement to consider more models than just the ubiquitous (and now relatively dated) BERT, we analyse four different pre-trained transformer models to fine-tune on our downstream NLU tasks.
\paragraph{BERT} \citep{devlin-etal-2019-bert} is arguably the most popular pre-trained large language model, the result of training a transformer over a very large amount of data for two relatively simple NLP tasks. BERT was designed to be adaptable to different prediction types -- e.g., regression, span prediction and, as in our case, sequence classification. We use the \texttt{base-uncased} version.
\paragraph{RoBERTa} \citep{liu2019roberta} improved on BERT by removing one of the pre-training tasks, modifying key hyperparameters and increasing the size of the training data. We use the \texttt{roberta-base} version.
\paragraph{ALBERT} \citep{Lan2020ALBERT:} managed to lower memory consumption and increase the training speed of BERT by using two specific parameter-reduction techniques. We use the \texttt{albert-base-v2} version.
\paragraph{DeBERTa} \citep{he2021deberta} improves the BERT and RoBERTa models using disentangled attention and enhanced mask decoder on half the size of its predecessors' training sets. We use version 3 where DeBERTa is further improved using ELECTRA-style pre-training. We use the \texttt{deberta-v3-small} configuration.

\subsection{Evaluation metrics}
We are primary interested in calibration performance, however we also check predictive performance drops that may occur as a result of the calibration step. Our calibration measure of choice is the expected calibration error, however Appendix \ref{sec:apx_logbrier} includes definitions and results for two additional measures: log loss and Brier score.

\paragraph{Expected Calibration Error}
To compute ECE, all predictions are grouped in $M$ bins of equal width, such that bin $B_m$ contains examples with confidence ranging in $(\frac{m-1}{M}, \frac{m}{M}]$.
ECE is defined as 
\[
\text{ECE}\coloneqq \frac{1}{n}\sum_{m=1}^M |B_m|\cdot |p(B_m)-\hat{p}(B_m)|
\]
where $p(B_m)$ is the true fraction of positive instances in bin $B_m$ and $\hat{p}(B_m)$ is the average estimated probability for predictions in bin $B_m$. For example, an ECE of 0.10 means that on average, the models' expected probability for a prediction is off by 10\%. It is important to note that ECE varies depending on the number of bins $M$: throughout our experiments we will report results for $M=10$.

\paragraph{$F_1$ score}
Macro-averaged $F_1$ score is defined as the arithmetic mean of the $F_1$ scores computed for each label. The $F_1$ score for a label $k$ is defined as 
\begin{equation}
    F_1^{(k)} = \frac{2P^{(k)}R^{(k)}}{P^{(k)} + R^{(k)}}
\end{equation}
where $P$ and $R$ are precision and recall.

\paragraph{Reliability bubble chart} A reliability diagram (see for example \citealp{niculescu2005predicting}) is a simple line plot that depicts the relationship between output probability and observed frequency (or accuracy, for the binary case). In this work we propose to replace the line plot with a bubble chart: the larger a bubble, the more examples have been assigned that particular probability by the model. Compared to the traditional reliability diagram, a bubble chart shows the model's preferences in terms of assigning probabilities, allowing for a better grasp of its sharpness.

\section{Results}
We report results for calibration and predictive accuracy. We include a final experiment where we try to reconstruct SST's original, real-valued labels from the corresponding binary labels alone.

\begin{table*}
\centering
\begin{tabular}{llrrrr}
\toprule
& & \textbf{QQP} & \textbf{BoolQ} & \textbf{CoLA} & \textbf{SST}\\
\midrule
\multirow{2}{5em}{ALBERT}   & \footnotesize\textsf{default} & 7.23   & 7.38   & 10.30  & 7.29 \\
                            & \footnotesize\textsf{IVAP}    & 0.52   & 3.32   & 3.14   & 3.38 \\
\midrule
\multirow{2}{5em}{BERT}     & \footnotesize\textsf{default} & 7.46   & 12.94  & 10.16 & 7.15 \\
                            & \footnotesize\textsf{IVAP}    & 0.44   & 3.35   & 3.09  & 2.76 \\
\midrule
\multirow{2}{5em}{DeBERTa}  & \footnotesize\textsf{default} & 6.18   & 10.79 & 10.25         & 4.20 \\
                            & \footnotesize\textsf{IVAP}    & 0.48   & 3.14  & 2.47 & 2.39 \\
\midrule
\multirow{2}{5em}{RoBERTa}  & \footnotesize\textsf{default} & 6.74   & 10.27          & 10.48 & 3.95 \\
                            & \footnotesize\textsf{IVAP}    & 0.49   & 2.79  & 2.92   & 2.99 \\
  
\bottomrule
\end{tabular}
\caption{
Expected calibration error (in \%) for default and IVAP models. A lower error means a lower discrepancy between estimated probability and model accuracy.
}
\label{tab:ece}
\end{table*}

\begin{figure}%[htbp]
\floatconts
    {fig:roberta-ivap-qqp}% label
    {\caption{Reliability bubble chart for the IVAP version of RoBERTa trained on the QQP dataset. The model is well-calibrated and the output probabilities are evenly distributed over the [0,1] interval (cfr. \figureref{fig:roberta-qqp}).}}% caption command
    {\includegraphics[width=0.6\textwidth]{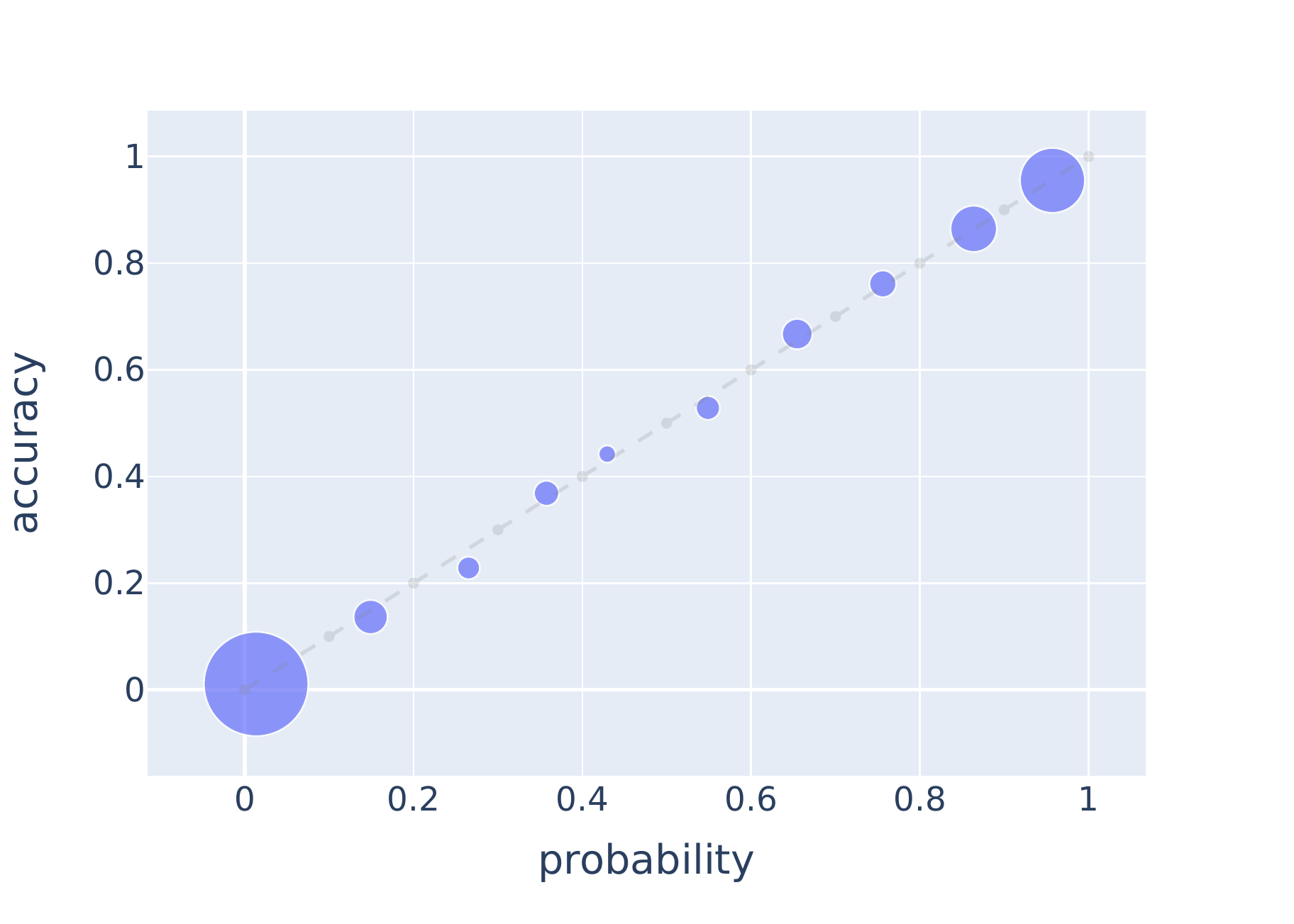}}
\end{figure}

\subsection{Calibration}
\tableref{tab:ece} shows the calibration performance of each transformer model along with its IVAP version. Each IVAP is a clear improvement over its original counterpart in terms of ECE. We notice how QQP, by far the largest of the 4 datasets, seems to attract better results: IVAPs are all almost perfectly calibrated ($\text{ECE}\sim0.005$). This may be due to the size of QQP's calibration set. As for the other three tasks, IVAPs still manage to reduce a model's ECE to 1/3 of the original, on average.

Among the four transformers, there is no dramatic difference between large (RoBERTa, DeBERTa) and smaller (BERT, ALBERT) models, even if the former actually manage to score better in 2 out of 4 tasks. Larger models however do benefit the most from being transformed into IVAP: the best calibration scores for IVAPs are shared between RoBERTa and DeBERTa.

For a more complete reporting of the calibration results, we include log loss and Brier scores in Appendix \ref{sec:apx_logbrier}.

By inspecting the reliability bubble charts, it is evident how IVAPs are sharper than their original counterparts: as \figureref{fig:roberta-ivap-qqp} shows, IVAP's probability scores are distributed relatively evenly over the [0,1] interval (compare with \figureref{fig:roberta-qqp}). 

As a final note, we report that IVAP calibration scores are more reliable, in the sense that they show a lower variability. For example, ECE for IVAPs never exceeds 5\% whereas it can vary from 1\% to 15\% among default models (see  \figureref{fig:ECEdist} in Appendix \ref{sec:apx_more}).

\subsection{Predictive accuracy}
Our main interest about predictive accuracy is to check if using IVAP -- and its reduced training set -- can harm the original model performance. \tableref{tab:f1} shows that there is a slight tendency of IVAP models to lose about 1\% $F_1$ score over their default version; however, this tendency wears off as the default model's performance increases, while sometimes IVAP can actually score even better (see RoBERTa on CoLA and SST).

In general, we note that while larger transformer models consistently achieve better scores than smaller ones, the gap in performance seem to become narrower for bigger datasets (see QQP).

Finally, the general trend is that more accurate models achieve better calibration (see \figureref{fig:ECEvsF1} in Appendix \ref{sec:apx_more}).

\begin{table*}
\centering
\begin{tabular}{llrrrr}
\toprule
& & \textbf{QQP} & \textbf{BoolQ} & \textbf{CoLA} & \textbf{SST}\\
\midrule
\multirow{2}{5em}{ALBERT}   & \footnotesize\textsf{default} & 0.90   & 0.70   & 0.79  & 0.87 \\
                            & \footnotesize\textsf{IVAP}    & 0.90   & 0.68   & 0.77   & 0.86 \\
\midrule
\multirow{2}{5em}{BERT}     & \footnotesize\textsf{default} & 0.90   & 0.69  & 0.80 & 0.87 \\
                            & \footnotesize\textsf{IVAP}    & 0.90   & 0.69   & 0.78  & 0.86 \\
\midrule
\multirow{2}{5em}{DeBERTa}  & \footnotesize\textsf{default} & 0.91   & 0.77 & 0.84  & 0.89 \\
                            & \footnotesize\textsf{IVAP}    & 0.91   & 0.76  & 0.83 & 0.89 \\
\midrule
\multirow{2}{5em}{RoBERTa}  & \footnotesize\textsf{default} & 0.91   & 0.77  & 0.81 & 0.89 \\
                            & \footnotesize\textsf{IVAP}    & 0.90   & 0.75  & 0.82   & 0.90 \\
  
\bottomrule
\end{tabular}
\caption{
Classification performance: $F_1$ scores for default and IVAP models.
}
\label{tab:f1}
\end{table*}

% \subsection{Examples of uncalibrated predictions}

\subsection{Estimating the degree of positive sentiment}
Some classification tasks are  ``more binary'' than others: label sets like $\mathcal{Y}=\{\text{alive}, \text{dead}\}$ define unambiguously a certain aspect of an object. However, binary labels may hide a more nuanced separation of the examples -- this is often the case of sentiment analysis. Because human sentiment is so subjective, it is hard to devise a labelling strategy that preserves the more subtle aspects of it, even (or maybe especially) in the simple case of $\mathcal{Y}=\{\text{negative, positive}\}$.

The Stanford Sentiment Treebank (SST, see Section \ref{datasets}), addresses this problem by assigning each example a real number $y\in[0,1]$ representing the \textit{degree} of positive sentiment of the sentence.\footnote{The final label is the average of the scores assigned by several human annotators.} In this work, we rounded those labels to the nearest integer to shape our task into binary sentiment analysis. This has already been done before, for example in the GLUE benchmark.

However, given the nature of the problem, and that on paper the aim of NLU should be ``understanding'' language, we think it would be preferable to build models that estimate uncertainty like humans do. Certainly, we can fit a simple regression model to predict SST's real-valued labels, but what if only binary labels were provided at training time? Would the probabilities generated by a model resemble the degree of human confidence in assigning a postive sentiment label?

As it turns out, the distribution of probability scores issued by NLU models doesn't really match that of confidence scores as judged by humans (see \figureref{fig:prob_dist} in Appendix \ref{sec:apx_more}). Nonetheless, IVAP models (and, we suspect, all well-calibrated and sharp models) manage to mimic human judgement in a better way. We compared the numeric labels in SST with the probabilities estimated by our transformer models and their IVAP variant. In \tableref{tab:r2} we show root mean squared error (RMSE) and $R^2$ score calculated over ground-truth and estimated degrees of positive sentiment. It is easy to verify how IVAPs always manage to achieve a better match of the human scores compared to their default counterparts.

\begin{table*}
\centering
\begin{tabular}{llrr}
\toprule
& & \textbf{RMSE} & \textbf{$R^2$}\\
\midrule
\multirow{2}{5em}{ALBERT}   & \footnotesize\textsf{default} & 0.28  & -0.22 \\
                            & \footnotesize\textsf{IVAP}    & 0.22  & 0.25 \\
\midrule
\multirow{2}{5em}{BERT}     & \footnotesize\textsf{default} & 0.29  & -0.27 \\
                            & \footnotesize\textsf{IVAP}    & 0.23  & 0.23 \\
\midrule
\multirow{2}{5em}{DeBERTa}  & \footnotesize\textsf{default} & 0.25  & 0.01 \\
                            & \footnotesize\textsf{IVAP}    & 0.23  & 0.20 \\
\midrule
\multirow{2}{5em}{RoBERTa}  & \footnotesize\textsf{default} & 0.26  & -0.05 \\
                            & \footnotesize\textsf{IVAP}    & 0.22  & 0.25\\
\bottomrule
\end{tabular}
\caption{
Estimation of the degree of positive sentiment in SST, when only binary labels are supplied at training time. Model's probabilities are compared to ground-truth labels in SST (lower RMSE and higher $R^2$ are better).
}
\label{tab:r2}
\end{table*}

\section{Conclusion}
We showed that Venn--ABERS predictors can be successfully applied to transformer models to obtain well-calibrated predictions for natural language understanding tasks. IVAPs were particularly effective when trained on a large dataset ($\text{ECE}<1\%$) and retained the classification accuracy of original transformer models. Moreover, IVAPs showed to be sharper: output probabilities were more evenly distributed in the $[0,1]$ interval and less condensed around a single value.

We restricted our experiments to the binary case, which is an obvious simplification of many real-world scenarios (e.g., detection of multiple intents in a chatbot, multiple topics of a message). However, \citet{pmlr-v60-manokhin17a} and \citet{pmlr-v152-johansson21a} both introduced methods to extend IVAPs to the multiclass case. This would allow us to directly compare Venn--ABERS to another calibration technique which is gaining traction in the deep learning community especially: temperature scaling \citep{pmlr-v70-guo17a}.

In terms of NLU, we avoided tasks like open question answering, machine translation and text summarization (all \textit{generative} tasks). Because there is not a single label $y$ for each example -- rather, a very large and potentially infinite set of possible labels -- an entire new and more useful definition for calibration may be needed.

The need for reliable NLU models will continue to grow as cutting-edge research is transformed into products for large audiences: users need to know when to trust a certain output. In a broader sense, we may say that a system with the ability of assessing its own uncertainty will always feel more ``intelligent'' than a blindly overconfident one. This reinforces the need for accurate calibration on the path towards a better AI.

\section*{Acknowledgements}
We would like to thank Prof. Alex Gammerman for his support and insightful suggestions. PG is in part supported by Centrica PLC.

\appendix

\section{Experimental setup}
\label{sec:apx_training}
All datasets except SST have hidden test sets as they are being used in ongoing competitions. For our experiments we concatenate their training and validation sets, shuffle the resulting dataset and split it again in training, validation and test set. For the IVAP training we further split each training set into 75\% proper training set and 15\% calibration set. The dataset sizes are summarized in \tableref{tab:splits}.

Because transformers are known to display some degree of variability in performance depending on the initial seed \citep{mosbach2021on}, we run 5 training trials and average their scores for all datasets, with the exception of QQP. In some occasions, a model would get stuck in a local minimum and perform extremely poorly -- this occurrences were removed from the calculation of the average as they would have skewed the result unnaturally.

All models were trained for 3 epochs with a learning rate of $2\cdot10^{-5}$ using the AdamW optimizer \citep{loshchilov2018decoupled}.

Training was performed on a single NVidia V100 hosted on the AWS platform. 

\begin{table*}
\centering
\begin{tabular}{lcccr}
\toprule
& \textbf{QQP} & \textbf{BoolQ} & \textbf{CoLA} & \textbf{SST}\\
\midrule
Train           & 323,416    & 9,427 & 7,468 & 8,544 \\
Validation      & 40,430     & 1,635 & 1,063 & 1,101 \\
Test            & 40,430     & 1,635 & 1,063 & 2,210 \\
\bottomrule
\end{tabular}
\caption{
Size of train, validation and test splits for the 4 datasets.
}
\label{tab:splits}
\end{table*}

\section{Additional calibration measures}
\label{sec:apx_logbrier}
We include results for calibration performance measured by log loss and Brier score, all averaged over 5 trials as detailed in Appendix \ref{sec:apx_training}.

Log loss, or cross-entropy loss, is based on how much a prediction with probability $p=\mathbb{P}(y=1)$ differs from the real label $y\in\{0,1\}$. It is defined as
\[
L(y,p)=-(y\log p + (1-y)\log(1-p))
\]
Intuitively, the lower the log loss averaged over the test set, the better a model is calibrated. Results of all models are summarised in \tableref{tab:logloss}.

Brier score is the mean squared error of the predictions over the test set, i.e.:
\[
L_B = \frac{1}{N}\sum_{i=1}^N (p_i-y_i)^2
\]
Unlike log loss, Brier score does not implode to $-\infty$ when a wrong prediction is given with $p=0$ or $p=1$. Results of all models are summarised in \tableref{tab:brier}.

\begin{table*}
\centering
\begin{tabular}{llrrrr}
\toprule
& & \textbf{QQP} & \textbf{BoolQ} & \textbf{CoLA} & \textbf{SST}\\
\midrule
\multirow{2}{5em}{ALBERT}   & \footnotesize\textsf{default} & 0.34   & 0.55   & 0.49  & 0.38 \\
                            & \footnotesize\textsf{IVAP}    & 0.24   & 0.54   & 0.41   & 0.32 \\
\midrule
\multirow{2}{5em}{BERT}     & \footnotesize\textsf{default} & 0.35   & 0.63  & 0.51 & 0.38 \\
                            & \footnotesize\textsf{IVAP}    & 0.23   & 0.55   & 0.40  & 0.32 \\
\midrule
\multirow{2}{5em}{DeBERTa}  & \footnotesize\textsf{default} & 0.29   & 0.53 & 0.49  & 0.29 \\
                            & \footnotesize\textsf{IVAP}    & 0.21   & 0.46  & 0.33 & 0.26 \\
\midrule
\multirow{2}{5em}{RoBERTa}  & \footnotesize\textsf{default} & 0.31   & 0.52  & 0.49 & 0.28 \\
                            & \footnotesize\textsf{IVAP}    & 0.22   & 0.47  & 0.36   & 0.27 \\
  
\bottomrule
\end{tabular}
\caption{
Log loss results for default and IVAP models.
}
\label{tab:logloss}
\end{table*}

\begin{table*}
\centering
\begin{tabular}{llrrrr}
\toprule
& & \textbf{QQP} & \textbf{BoolQ} & \textbf{CoLA} & \textbf{SST}\\
\midrule
\multirow{2}{5em}{ALBERT}   & \footnotesize\textsf{default} & 0.082   & 0.182  & 0.137  & 0.103 \\
                            & \footnotesize\textsf{IVAP}    & 0.072   & 0.182  & 0.127  & 0.099 \\
\midrule
\multirow{2}{5em}{BERT}     & \footnotesize\textsf{default} & 0.081   & 0.206  & 0.129  & 0.107 \\
                            & \footnotesize\textsf{IVAP}    & 0.068   & 0.184  & 0.122  & 0.097 \\
\midrule
\multirow{2}{5em}{DeBERTa}  & \footnotesize\textsf{default} & 0.072   & 0.161  & 0.113  & 0.081 \\
                            & \footnotesize\textsf{IVAP}    & 0.064   & 0.150  & 0.098  & 0.078 \\
\midrule
\multirow{2}{5em}{RoBERTa}  & \footnotesize\textsf{default} & 0.076   & 0.159  & 0.123  & 0.082 \\
                            & \footnotesize\textsf{IVAP}    & 0.067   & 0.154  & 0.109  & 0.078 \\
  
\bottomrule
\end{tabular}
\caption{
Brier scores for default and IVAP models.
}
\label{tab:brier}
\end{table*}

\section{Additional plots}
\label{sec:apx_more}
\figureref{fig:ECEdist} shows how IVAP models are more consistently calibrated regardless of the initial seed used for model generation. On the other hand, default models can be more or less well-calibrated, depending on how ``lucky'' the initial seed is.

\figureref{fig:ECEvsF1} shows that more accurate models tend to be better calibrated. However, this tendency is attenuated when using IVAP models.

\figureref{fig:prob_dist} shows the distribution of human scores in the SST dataset, together with IVAP's and the default model's estimations. IVAP tries to recreate the essentially bimodal distribution of human scores, while the default model (in this case a fine-tuned DeBERTa) struggles to do so and prefers extreme values of probability.

\begin{figure}%[htbp]
\floatconts
    {fig:ECEdist}% label
    {\caption{Distribution of expected calibration errors over all datasets, models and trials.}}% caption command
    {\includegraphics[width=0.6\textwidth]{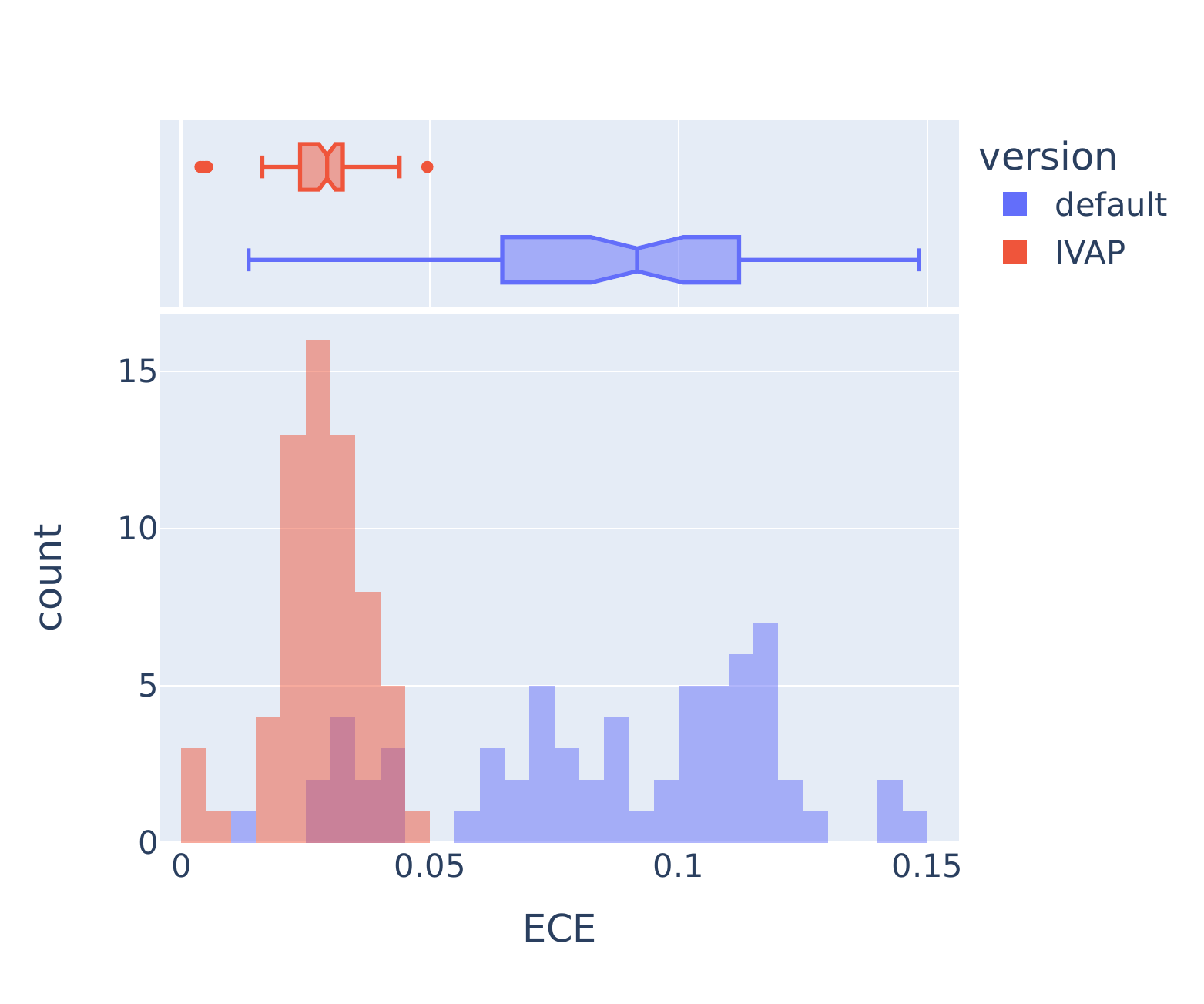}}
\end{figure}

\begin{figure}%[htbp]
\floatconts
    {fig:ECEvsF1}% label
    {\caption{Trend of expected calibration error versus $F_1$ score for all models and datasets.}}% caption command
    {\includegraphics[width=0.6\textwidth]{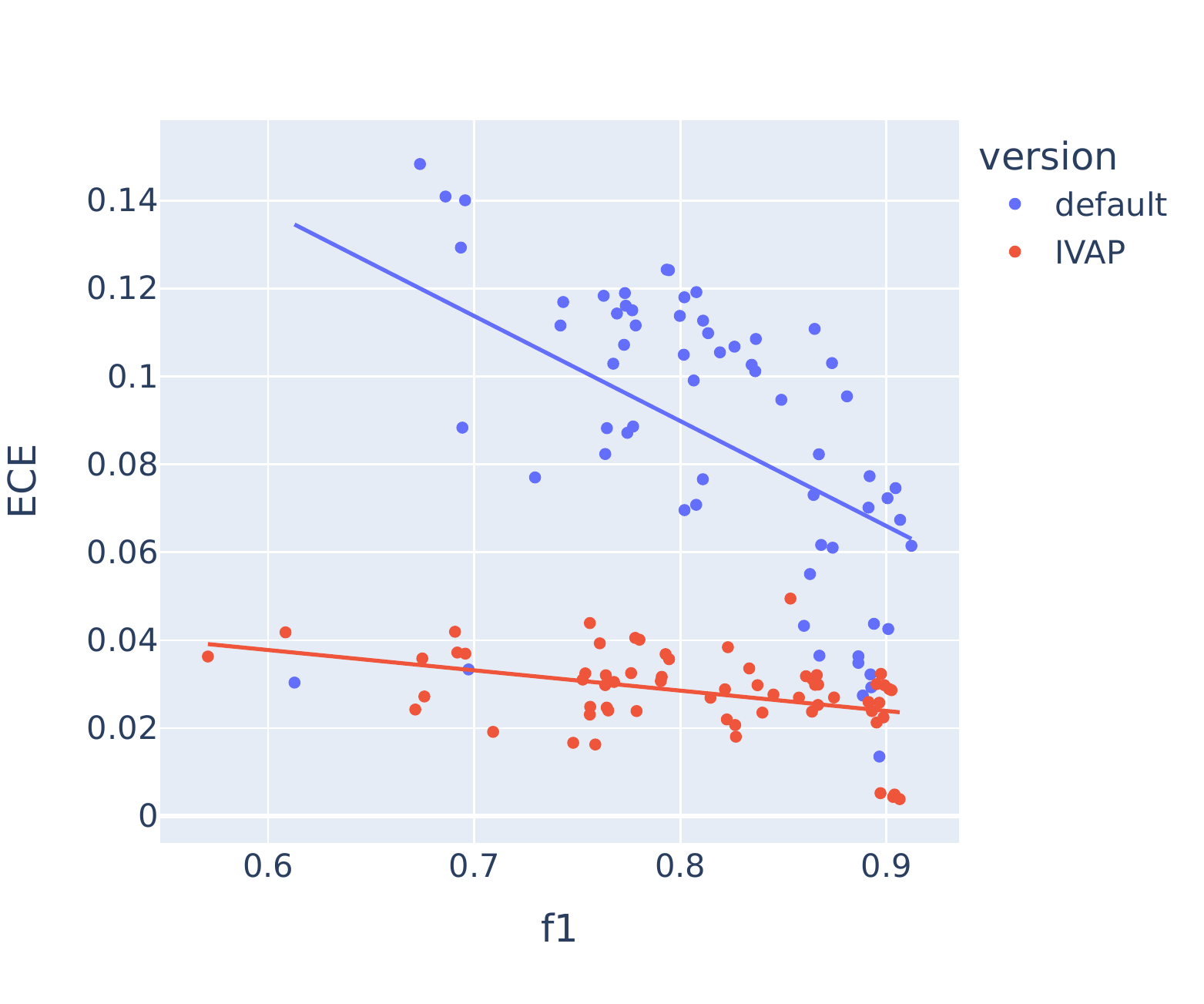}}
\end{figure}

\begin{figure}%[htbp]
\floatconts
    {fig:prob_dist}% label
    {\caption{Distribution of ground-truth scores in SST, compared to IVAP's and default model's estimations.}}% caption command
    {\includegraphics[width=0.6\textwidth]{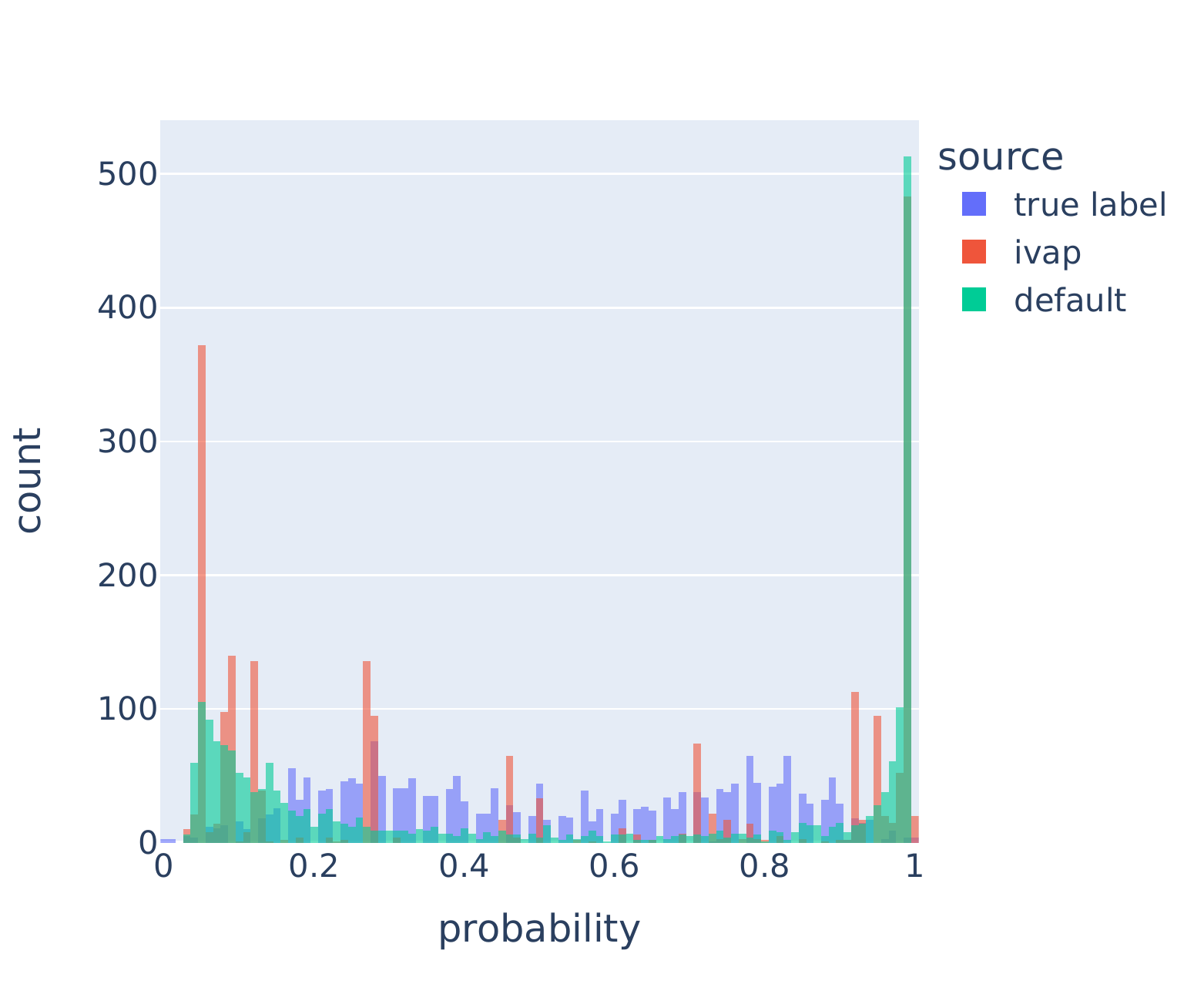}}
\end{figure}

\bibliography{anthology}

\end{document}